\documentclass[sigconf]{acmart}

\AtBeginDocument{%
  }

\setcopyright{rightsretained}
\copyrightyear{2023}
\acmYear{2023}
\acmDOI{10.1145/3577193.3593704}
\acmConference[ICS '23]{2023 International Conference on Supercomputing}{June 21--23, 2023}{Orlando, FL, USA}
\acmBooktitle{2023 International Conference on Supercomputing (ICS '23), June 21--23, 2023, Orlando, FL, USA}
\acmPrice{15.00}
\acmISBN{979-8-4007-0056-9/23/06}

\usepackage{listings}
\usepackage{xspace}
\usepackage{subfig}
\usepackage[export]{adjustbox}
\usepackage{todonotes}
\usepackage{tablefootnote}

\usepackage{booktabs}
\usepackage{multirow}
\usepackage{threeparttable}
\usepackage{tikz}
\usepackage{balance}

\newcommand{\circled}[1]{\tikz[baseline=(char.base)]{
            \node[circle,fill=.,inner sep=0.4pt] (char) {\textcolor{white}{#1}};}}

\definecolor{dkgreen}{RGB}{0,64,0}
\definecolor{ltgray}{RGB}{245,245,245}
\definecolor{mauve}{RGB}{139,0,139}

\begin{document}

\title{A Hybrid Tensor-Expert-Data Parallelism Approach to Optimize Mixture-of-Experts Training}

\author{Siddharth Singh}
\email{ssingh37@umd.edu}
\orcid{0000-0002-2756-4290}
\affiliation{%
  \institution{Department of Computer Science, University of Maryland}
  \city{College Park}
  \state{Maryland}
  \country{USA}
  \postcode{20742}
}

\author{Olatunji Ruwase}
\email{olruwase@microsoft.com}
\orcid{0000-0002-5508-0728}
\affiliation{%
  \institution{Microsoft, Inc.}
  \city{Redmond}
  \state{Washington}
  \country{USA}
}

\author{Ammar Ahmad Awan}
\email{ammar.awan@microsoft.com}
\orcid{0000-0002-6272-3760}
\affiliation{%
  \institution{Microsoft, Inc.}
  \city{Redmond}
  \state{Washington}
  \country{USA}
}

\author{Samyam Rajbhandari}
\email{samyamr@microsoft.com}
\orcid{0000-0002-0386-8759}
\affiliation{%
  \institution{Microsoft, Inc.}
  \city{Redmond}
  \state{Washington}
  \country{USA}
}

\author{Yuxiong He}
\email{yuxhe@microsoft.com}
\orcid{0000-0003-0478-8854}
\affiliation{%
  \institution{Microsoft, Inc.}
  \city{Redmond}
  \state{Washington}
  \country{USA}
}

\author{Abhinav Bhatele}
\email{bhatele@cs.umd.edu}
\orcid{0000-0003-3069-3701}
\affiliation{%
  \institution{Department of Computer Science, University of Maryland}
  \city{College Park}
  \state{Maryland}
  \country{USA}
  \postcode{20742}
}

\renewcommand{\shortauthors}{Singh et al.}

\ccsdesc[500]{Computing methodologies~Massively parallel algorithms}
\ccsdesc[500]{Computing methodologies~Natural language generation}

\begin{abstract}
Mixture-of-Experts (MoE) is a neural network architecture that adds sparsely
activated expert blocks to a base model, increasing the number of parameters
without impacting computational costs. However, current distributed deep
learning frameworks are limited in their ability to train high-quality MoE
models with large base models.  In this work, we present DeepSpeed-TED, a
novel, three-dimensional, hybrid parallel algorithm that combines data, tensor,
and expert parallelism to enable the training of MoE models with 4--8$\times$
larger base models than the current state-of-the-art. We also describe memory
optimizations in the optimizer step, and communication optimizations that
eliminate unnecessary data movement. We implement our approach in DeepSpeed and
achieve speedups of 26\% over a baseline (i.e. without our communication
optimizations) when training a 40 billion parameter MoE model (6.7 billion base
model with 16 experts) on 128 V100 GPUs.

\end{abstract}

\keywords{Parallel Deep Learning, Mixture-of-Experts, Tensor Parallelism, Expert Parallelism}

\maketitle

\section{Introduction}
\begin{figure}[h]
    \centering
      \includegraphics[width=\columnwidth]{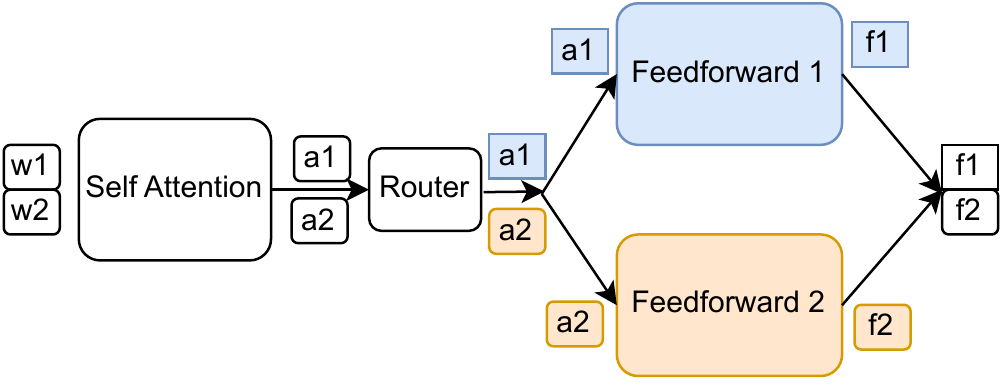}
    \caption{A single Mixture-of-Experts (MoE) layer with two ``experts'' or
  feedforward blocks. The input batch has two tokens, w1 and w2. We use the prefixes `w', `a', 
  and `f' to denote the input activations to the layer, output activations of self-attention 
  and feedforward blocks respectively. Similarly we label each activation with an integer suffix 
  corresponding to its token. Note that each token is uniquely routed to a single expert by a parameterized routing 
  function.}
    \label{fig:moe-schematic}
\end{figure}

Contemporary state-of-the-art AI algorithms have come to rely on neural
networks such as GPT-3~\cite{gpt-3} and MT-NLG~\cite{megatron-turing-nlg-530b}
with hundreds of billion of parameters. However, training or running inference
on models of this size has become prohibitively expensive due to their
significantly large computational costs. To alleviate this issue, deep learning
researchers have turned their attention to the Mixture-of-Experts (MoE)
architecture~\cite{og-moe, switch-transformer, g-shard}, which offers a way to
increasing the parameter count of a model without increasing computational
costs.  MoE models augment the layers of a vanilla
transformer~\cite{transformer} model (called the base model in MoE parlance)
with multiple ``experts'' or feedforward blocks and a parameterized routing
function that uniquely maps each input token to a unique expert.
Figure~\ref{fig:moe-schematic} illustrates the forward pass of a single MoE
layer with two experts and an input batch of two tokens. Since each token is
only processed by one expert, the effective computation cost per token (and
thus the total training cost) remains fixed (in comparison to the base model)
and is independent of the number of experts. 

Unfortunately, there is a limit to the improvements in model quality that can
be achieved by simply increasing the number of experts~\cite{og-moe}. For
example, in an experiment that studied the effect of adding experts to the T5
architecture~\cite{t5-transformer} as the base model, Fedus et al.~observed
diminishing improvements in the test set accuracy beyond 64-128
experts~\cite{switch-transformer}. In fact, for training high quality
Mixture-of-Experts models, it is imperative that the base models' size (number
of parameters) is increased along with the number of experts~\cite{ds-moe-ml}.  

In this regard, current state-of-the-art distributed deep learning frameworks
are inadequate for training such MoEs with large base models. They either
support limited-sized base models or use inefficient parallel algorithms that
lead to high communication costs. Hence, it is crucial to develop a distributed
framework that can support the training of MoEs with large base models on
multi-GPU clusters and do so efficiently, while keeping communication costs
low. In this work, we present a three-dimensional, hybrid parallel framework,
DeepSpeed-TED, that combines ZeRO's data parallelism~\cite{sc2020zero},
MegatronLM's tensor parallelism~\cite{megatronlm}, and DeepSpeed-MoE's expert
parallelism~\cite{ds-moe-systems} to train MoE models that are built using
extremely large base models. We demonstrate how the combination of these three
dimensions of parallelism  allows our framework to train 4--8 $\times$ larger
base models compared to DeepSpeed-MoE~\cite{ds-moe-systems}, a state-of-the-art
parallel framework that employs only two of these dimensions (data and expert).
To the best of our knowledge, this is the first effort that combines these
three state-of-the-art parallel deep learning algorithms for training MoEs on
multi-GPU clusters.

We identify and resolve two bottlenecks that emerge with a naive combination of
these three forms of parallelism.  The first is a significant increase in
memory usage in the optimizer, which limits the base model sizes supported by
our hybrid parallel approach. To alleviate this issue, we propose a tiled
version of the optimizer that processes model parameters in groups (or tiles)
of fixed size, and decreases peak memory consumption by reusing GPU memory
across the parameter tiles.  The second bottleneck is related to communication
costs, where a considerable amount of training time is spent in collective
communication pertaining to expert and tensor parallelism.  We identify two
regions in this hybrid training procedure where messages are communicated
unnecessarily among the worker GPUs, and propose novel communication
optimizations that resolve this problem.  For a 40 billion parameter MoE (6.7
billion base model with 16 experts) on 128 V100 GPUs of Summit, our
optimizations reduce the overall collective communication time by 42\% and lead
to a significant improvement of 26\% in the training time. DeepSpeed-TED is
open source, and has been integrated in
DeepSpeed\footnote{\url{https://github.com/microsoft/DeepSpeed}}, a
state-of-the-art distributed deep learning framework.

The main contributions of this paper are as follows:
\begin{itemize}
\item A highly scalable first of its kind three-dimensional, hybrid parallel
framework that combines ZeRO's data~\cite{sc2020zero}, Megatron-LM's
tensor~\cite{megatronlm}, and DeepSpeed-MoE's expert~\cite{ds-moe-systems}
parallelism to enable the training of Mixture-of-Experts with large base
models. 
\item A tiled version of an optimizer that alleviates a significant memory
spike in the optimizer step that arises from combining the three aforementioned
forms of parallelism.
\item Communication optimizations that eliminate unnecessary communication in
our hybrid parallel algorithm which significantly reduce  collective
communication times.
\end{itemize}

\section{Background}
\label{sec:bg}
In this section, we provide a background on Mixture-of-Experts (MoE), and the
three forms of parallelism used in this work -- tensor
parallelism~\cite{megatronlm}, expert parallelism~\cite{ds-moe-systems}, and
data parallelism~\cite{sc2020zero}. 

\subsection{Mixture-of-Experts}
\label{bg:moe}

Proposed by Shazeer et al.~\cite{og-moe} in 2017, Mixture-of-Experts (MoEs) are
a family of neural network architectures with an interesting property that
their parameter set can be made arbitrarily large without increasing their
computational costs.  This is achieved by adding sparsely activated expert
blocks to the layers of a dense neural network (called the base model).  A
parameterized routing function is added before these expert blocks that maps
its input tokens to a unique expert.  Since each token is computed upon by only
one expert, the total computation cost of training remains fixed (same as the
base model) and is independent of the number of experts. MoEs thus offer a
unique way to increase the number of parameters of a given base model, and thus
its performance on any task~\cite{scaling-laws}, without any increase in
computational costs. Although Shazeer et al.~used LSTMs~\cite{lstm-paper} as
their base models, contemporary work on MoEs largely employ the transformer
architecture~\cite{transformer} as the base models~\cite{switch-transformer,
g-shard, ds-moe-ml, taming-moe, wider-moe, stable-moe, glam-moe, vision-moe,
fb-moe}. 

\subsection{Data Parallelism and ZeRO}
\label{bg:dp-zero}

Under data parallelism, worker GPUs house a copy of the neural network and work
on mutually exclusive shards of the input batch.  After the backward pass they
synchronize their local gradients via an all-reduce function call. However, a
major limitation of data parallelism is that each GPU has to have enough memory
to store the entire parameter set of a neural network, and also its gradients
and optimizer states.  To resolve this issue, Rajbhandari et al.~proposed the
Zero Redundancy Optimizer or ZeRO which is aimed at eliminating this redundant
memory consumption in data parallel GPUs~\cite{sc2020zero}. Their method has
three stages, which progressively save more memory albeit at the cost of
increased communication. In this work, we consider the first stage of their
optimization which only shards the optimizer states across data parallel ranks.

\subsection{Expert Parallelism and DeepSpeed-MoE}

After routing, the computation of an expert block of in an MoE layer is
independent of other experts. Expert parallelism exploits this property by
placing unique expert blocks on each GPU and computing them in an
embarrassingly parallel fashion. Tokens are mapped to their corresponding
experts by all-to-all communication within the participating GPUs.  Due to its
simplicity and effectiveness, expert parallelism is used in many parallel
frameworks for training or running inference on MoEs \cite{hetu-moe, se-moe,
switch-transformer, g-shard, fb-moe}. In this work, we use DeepSpeed-MoE's
implementation of expert parallelism~\cite{ds-moe-systems}.

\subsection{Tensor Parallelism and Megatron-LM}

Tensor parallelism involves partitioning the computation of a neural network
layer across GPUs.  Shoeybi et al.~introduce MegatronLM, a tensor parallel
algorithm to parallelize the computation of layers in a transformer neural
network~\cite{megatronlm}. Their method is aimed at parallelizing a pair of
consecutive fully-connected layers, which are found in the self-attention and
feedforward blocks of the transformer~\cite{transformer}.  Their algorithm has
seen significant adoption for training many large language models like
Megatron-Turing NLG 530B \cite{megatron-turing-nlg-530b}, Bloom-176 B
\cite{bloom176b}, Turing NLG \cite{sc2020zero} etc.

\section{TED: A Hybrid Tensor-Expert-Data Parallel Approach}
\label{sec:method}
By adding sparsely activated experts, the Mixture-of-Experts architecture
allows us to make a given neural network, i.e. the base model, arbitrarily
large while keeping its computation cost unchanged.  However, merely increasing
the number of experts yields diminishing returns in model generalization beyond
64--128 experts~\cite{switch-transformer}. To build high quality MoEs, it is
imperative that we increase the base model sizes as well as the number of
experts~\cite{ds-moe-ml}. In this section, we provide an overview of TED, our
hybrid parallel approach which combines DeepSpeed-MoE's
expert~\cite{ds-moe-systems}, MegatronLM's tensor~\cite{megatronlm} and ZeRO's
data~\cite{sc2020zero} parallelism, to enable the training of such MoEs with
extremely large multi-billion parameter base models on multi-GPU clusters. In
this work, we use the first stage of ZeRO, which shards the optimizer states
across data parallel GPUs. While further stages of their optimizations
(stage-2, 3, offload~\cite{zero_offload} and infinity~\cite{zero_infinity}) can
support training of larger models, this happens at a cost to performance. 

\begin{figure}[h]
    \centering
      \includegraphics[width=\columnwidth]{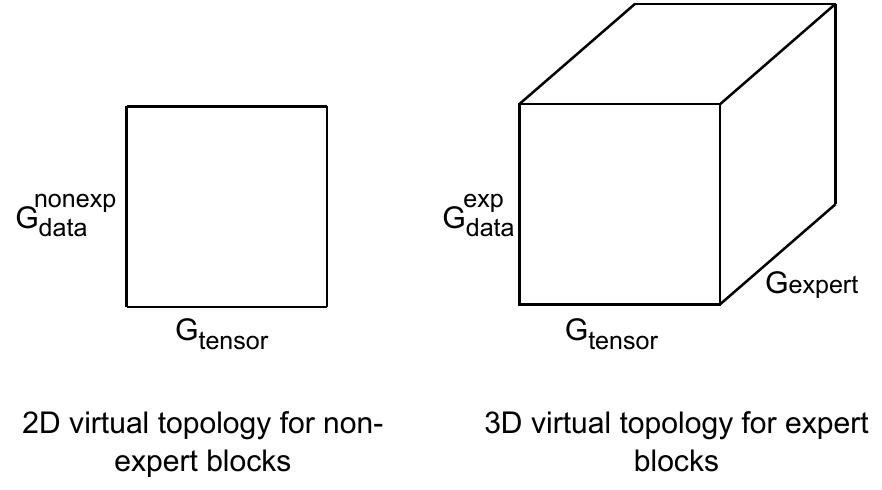}
    \caption{TED uses a two dimensional hybrid of tensor and data parallelism to parallelize the computation of non-expert blocks. 
    Whereas, it utilizes all three of tensor, expert, and data parallelism to parallelize expert blocks.}
\label{fig:topology}
\end{figure}

We use the terms non-expert and expert blocks interchangeably with
self-attention and feedforward blocks respectively.  Note that TED parallelizes
the computation of expert and non expert blocks in a different manner. This is
because expert parallelism is only applicable to the feedforward blocks of the
transformer base model.  Thus, TED uses a two dimensional hybrid of tensor and
data parallelism to parallelize the non-expert blocks.  Whereas, it utilizes
all three of tensor, expert, and data parallelism for the expert blocks. Under
TED, we organize available GPUs into two different virtual topologies for the
non-expert and expert blocks. We illustrate these topologies in
Figure~\ref{fig:topology}.

For the non-expert blocks, we maintain a two dimensional (2D) topology of GPUs,
one dimension each for tensor and data parallelism. In this topology, GPUs in a
row implement tensor parallelism, and we refer to a row of GPUs as a tensor
parallel group. Similarly, TED realizes data parallelism across columns of
GPUs, and we refer to these columns as data parallel groups.  Likewise, for the
expert blocks, we maintain a three dimensional (3D) topology of GPUs, one each
for tensor, expert, and data parallelism. To form the tensor parallel groups
for the expert blocks, we reuse the tensor parallel groups formed in the 2D
topology for the non-expert blocks. However, we further decompose the data
parallel groups of the non-expert blocks into a 2D topology to form groups for
expert parallelism and data parallelism for the expert blocks. We define
$\mathit{G_{tensor}}$ and $\mathit{G^{nonexp}_{data}}$ as the size of the
tensor parallel and non-expert data parallel groups respectively. Similarly, we
define $\mathit{G_{expert}}$ and $\mathit{G^{exp}_{data}}$ as the size of the
expert parallel and expert data parallel groups respectively. Following prior
work~\cite{sc2020zero}, we always set $\mathit{G_{expert}}$ to the number of
experts in the model for performance considerations. Note that given a number
GPUs, $\mathit{G}$, the following relation always holds true:
\begin{equation}
    \mathit{{G}_{tensor}}\times\mathit{G_{expert}}\times\mathit{G^{exp}_{data}}=\mathit{{G}_{tensor}}\times\mathit{G^{nonexp}_{data}}=G \label{eqn:relation}
\end{equation}

\begin{figure*}[h]
    \centering
      \includegraphics[width=2\columnwidth]{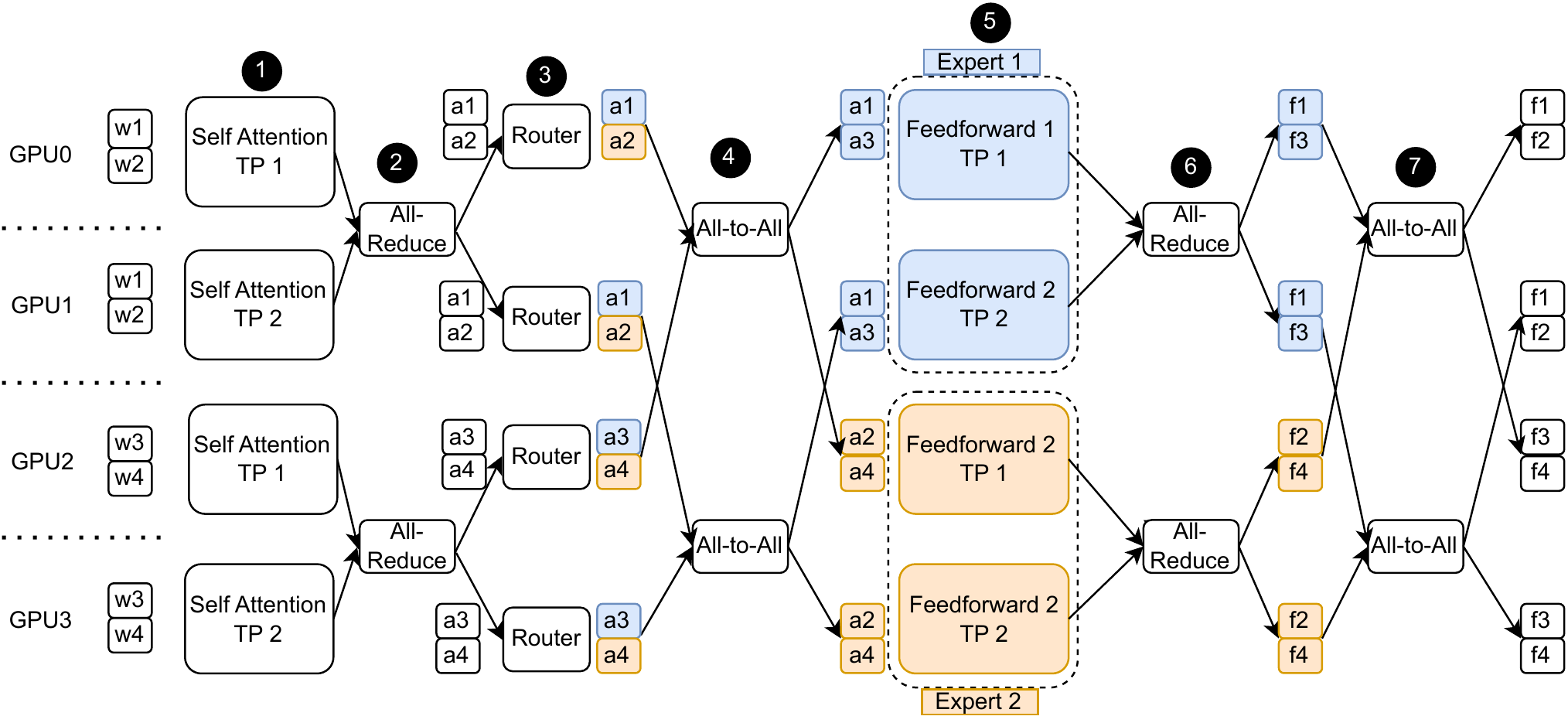}
    \caption{Forward pass of an MoE layer with two experts on four GPUs using TED. We use a $\mathit{{G}_{tensor}}\times\mathit{G^{nonexp}_{data}}=2\times2$ topology 
    for the non-expert self-attention blocks and $\mathit{{G}_{tensor}}\times\mathit{{G}_{expert}}\times\mathit{G^{exp}_{data}}=2\times2\times1$ topology 
    for the expert feedforward blocks. We use the prefixes `w', `a', and `f' to denote the input activations to the layer,
  output activations of self-attention and feedforward blocks respectively. 
  Similarly we label each activation with an integer suffix corresponding to its token.
  Suffixes TP 1 and TP 2 denote the two tensor parallel partitions of the attention and 
  feedforward blocks. The input batch consists of four tokens, with tokens 1 and 3 routed to the
  first expert (colored blue), and tokens 2 and 4 routed to the second expert (colored
  yellow).}
    \label{fig:tp-ep-dp}
  \end{figure*}

In Figure~\ref{fig:tp-ep-dp}, we illustrate the forward pass of an MoE layer
with two experts on four GPUs. As mentioned previously, we set
$\mathit{G_{expert}}$ to the number of experts i.e. 2. The other degrees of
parallelism are $\mathit{{G}_{tensor}}=2$, $\mathit{G^{nonexp}_{data}}=2$,
$\mathit{G_{expert}}=2$, and $\mathit{G^{exp}_{data}}=1$. We partition the
parameters of the self-attention block (non-expert) and the two feed forward
blocks (experts) as per the semantics of MegatronLM's tensor parallelism and
place the first partition on GPUs 0 and 2 and the second partition on GPUs 1
and 3. GPUs (0,1) and (2,3) thus form the two tensor parallel groups. GPU pairs
(0,2) and (1,3) comprise the data parallel groups for the non-expert
parameters. The same GPU pairs however comprise the expert parallel groups for
the expert parameters.  The four GPUs individually form singleton data parallel
groups for the expert parameters. 

Let us now discuss how our hybrid parallel algorithm computes the forward pass
of an MoE layer. As an example, we use an input batch with four tokens
(numbered 1-4) in Figure~\ref{fig:tp-ep-dp}. The tensor parallel group of GPUs
0 and 1 compute on tokens 1 and 2, whereas the tensor parallel group of GPUs 2
and 3 compute on tokens 3 and 4. Each GPU first computes their partition of the
self-attention block (\circled{1}) and then issues an all-reduce (\circled{2})
to aggregate the complete output activations (prefixed by 'a') for their
respective tokens.  Now, each GPU applies the MoE routing function to their
local tokens (\circled{3}).  We assume that the routing function maps tokens 1
and 3 to the first expert i.e. feedforward 1, and tokens 2 and 4 to the second
expert i.e. Feedforward 2. (\circled{4}) Now, an all-to-all communication
primitive is issued in expert parallel groups to route the tokens as per the
mapping decided by the routing function.  Let us look at the expert parallel
group of GPUs 0 and 2 to understand this all-to-all communication call.  On GPU
0, token 1 has been mapped to the first expert and token 2 has been mapped to
the second expert. Therefore, we want to retain a1 and send a2 to GPU 2 which
houses the second expert. Similarly, on GPU 2, we want to retain a4 and send a3
over to GPU 0. Note that this communication pattern matches the semantics of an
all-to-all communication primitive exactly. After the all-to-all has completed
each GPU computes their tensor-parallel partitions of the expert feed forward
blocks (\circled{5}) and issue an all-reduce to aggregate the complete output
(\circled{6}). The final all-to-all communication call in the expert parallel
groups (\circled{7}) essentially inverts the first all-to-all (\circled{4}) and
brings back the tokens to their original GPUs. This is how our three
dimensional hybrid parallel approach computes the forward pass of an MoE layer. 

During the backward pass computation proceeds in the reverse direction i.e.
(\circled{7} - \circled{1}). The all-to-all communication at \circled{7} and
\circled{4} calls are reversed. For example, consider \circled{7}, wherein the
input to the all-to-all on GPU 0 would be gradients of the loss w.r.t.~f1 and
f2. Similarly for GPU 2, it would be the gradients w.r.t.~ f3 and f4. Now,
after the all-to-all, the outputs on GPU 0 would be gradients of the loss
w.r.t~f1 and f3, and on GPU 2 it would be gradients of the loss w.r.t.~f2 and
f4. The all-reduce function calls (\circled{4}, \circled{6}) are applied to the
gradients w.r.t the input activations instead of the output. For more details
about this all-reduce call, we refer the reader to Narayanan et
al.~\cite{megatronlm-2}. Note that total amount of communication i.e. two
all-reduces and two all-to-alls is the same as that of the forward pass.
Finally, the data parallel groups synchronize their gradients via another
all-reduce call, which completes the backward pass.

\subsection{A Model for Memory Consumption}

We now derive the extent to which TED can increase the base model sizes as
compared to prior work like DeepSpeed-MoE~\cite{ds-moe-systems}, which only
employ data and expert parallelism. Following previous work, we assume that
every alternate layer has expert feedforward modules~\cite{g-shard,
switch-transformer, ds-moe-ml}.  Let $\mathit{NP_{base}}$ denote the number of
parameters in the base model and $E$ denote the number of experts. Let $G$ be
the number of GPUs. Note that two-thirds of the parameters in the base model
reside in feed-forward blocks, and the remaining one-third in self-attention
blocks~\cite{megatronlm-2}. Since only half of the feedforward blocks are
designated as experts, the total number of expert parameters,
$\mathit{NP_{exp}}$, in an MoE model are:
\begin{equation} 
     \mathit{NP_{exp}} = E \times \frac{1}{2} \times \left( \frac{2}{3} \times \mathit{NP_{base}} \right) =\frac{E}{3} \times \mathit{NP_{base}}  \label{eq:pe}
\end{equation}

Now, the non-expert parameters are comprised of parameters in all the
self-attention blocks and half of the feed-forward blocks. Thus, the total
number of non-expert parameters, $\mathit{NP_{nonexp}}$, is
\begin{equation} 
    \mathit{NP_{nonexp}} = \frac{1}{2} \times \left( \frac{2}{3} \times \mathit{NP_{base}} \right) + \frac{1}{3} \times \mathit{NP_{base}} =\frac{2}{3} \times \mathit{NP_{base}}  \label{eq:pne}
\end{equation}

Rajbhandari et al.~\cite{sc2020zero} prove that the lower bound of memory
consumption per GPU with ZeRO stage-1  is $\left( 4 +
\frac{12}{\mathit{G_{data}}} \right) \times \mathit{NP_{gpu}}$, where
$\mathit{G_{data}}$ is the degree of data parallelism and $\mathit{NP_{gpu}}$
is the number of parameters of the model per GPU. Now, we use this formulation
to derive a lower bound on memory consumption per GPU for TED as follows:
\begin{equation}
    \mathit{M_{gpu}} \geq \left( 4 + \frac{12}{\mathit{G^{nonexp}_{data}}} \right) \times \mathit{NP^{nonexp}_{gpu}} + 
    \left( 4 + \frac{12}{\mathit{G^{exp}_{data}}} \right) \times \mathit{NP^{exp}_{gpu}} \label{eqn:zero}
\end{equation}

Here, $\mathit{NP^{nonexp}_{gpu}}$ and $\mathit{NP^{exp}_{gpu}}$ are the number
of expert and non expert parameters per GPU.  As discussed previously,
$\mathit{G^{nonexp}_{data}}$ and $\mathit{G^{exp}_{data}}$ are the degrees of
data parallelism for the non-expert and expert blocks respectively. Now, let us
try to derive the values of $\mathit{NP^{nonexp}_{gpu}}$ and
$\mathit{NP^{exp}_{gpu}}$, starting with the former. MegatronLM's tensor
parallelism divides the parameters of a model equally among the GPUs in a
tensor parallel group.  Since the size of a tensor parallel group in TED is
$\mathit{G_{tensor}}$, we can write $\mathit{NP^{nonexp}_{gpu}} =
\frac{\mathit{NP_{nonexp}}}{\mathit{G_{tensor}}}$. However, the expert
parameters are divided within both the tensor parallel and expert parallel
groups. As discussed previously, we use a degree of expert parallelism equal to
the number of experts i.e. $\mathit{G_{expert}}=E$. Thus,
$\mathit{NP^{exp}_{gpu}} = \frac{\mathit{NP_{exp}}}{\mathit{G_{tensor}}\times
E}$.  Also, it follows from Equation~\ref{eqn:relation} that
$\mathit{G^{nonexp}_{data}}=\frac{\mathit{G}}{\mathit{G_{tensor}}}$ and
$\mathit{G^{exp}_{data}}=\frac{\mathit{G}}{\mathit{G_{tensor}}\times\mathit{G_{expert}}}=\frac{\mathit{G}}{\mathit{G_{tensor}}\times\mathit{E}}$.
Substituting these values in Equation~\ref{eqn:zero}, we get 
\begin{align}
    \mathit{M_{gpu}} &\geq \left( 4 + \frac{12\mathit{G_{tensor}}}{G} \right) \times \frac{\mathit{NP_{nonexp}}}{\mathit{G_{tensor}}} +
    \left( 4 + \frac{12\mathit{G_{tensor}} E }{G} \right) \times \frac{\mathit{NP_{exp}}}{\mathit{G_{tensor}}E} \nonumber \\
    &\geq \frac{4}{\mathit{G_{tensor}}} \left( \mathit{NP_{nonexp}} + \frac{\mathit{NP_{exp}}}{E}\right) 
    + \frac{12}{G} (\mathit{NP_{nonexp}} + \mathit{NP_{exp}}) \nonumber 
\end{align}

Now substituting from Equation~\ref{eq:pe} and~\ref{eq:pne}, we get 
\begin{align}
    \mathit{M_{gpu}} &\geq \frac{4}{\mathit{G_{tensor}}} \left( \frac{2}{3} \mathit{NP_{base}} + \frac{\mathit{NP_{base}}}{3} \right) 
    + \frac{12}{G} (\frac{2}{3} \mathit{NP_{base}} + \frac{E}{3} \mathit{NP_{base}}) \nonumber \\
    &\geq \frac{4\mathit{NP_{base}}}{\mathit{G_{tensor}}} +  \frac{4(E+2)}{G}  \mathit{NP_{base}} \nonumber \\
    &\geq 4\mathit{NP_{base}} \times \left(\frac{1}{\mathit{G_{tensor}}} + \frac{E+2}{G} \right) \label{eqn:lower-bound-full}
\end{align}

Equation~\ref{eqn:lower-bound-full} can be used to derive an upper bound on the
largest possible base model size that our framework can train, given enough
number of GPUs. Note that as we increase the number of GPUs involved in
training, the second term becomes negligible compared to the first. This gives
us,
\begin{align}
    \mathit{M_{gpu}} &\geq \frac{4\mathit{NP_{base}}}{\mathit{G_{tensor}}} \nonumber \\
    \implies \mathit{NP_{base}} &\leq \frac{\mathit{G_{tensor}}}{4} \times \mathit{M_{gpu}} \label{eqn:upper-bound-base}
\end{align}

Note that substituting $\mathit{G_{tensor}}=1$ in
Equation~\ref{eqn:upper-bound-base} gives us the base model upper bound for
Rajbhandari et al.~\cite{ds-moe-systems}, the current state-of-the-art for
training MoEs. Thus, we have shown that our system enables the training of
$\mathit{G_{tensor}}\times$ larger base models compared to the previous
state-of-the-art. Note that the maximum degree of tensor parallelism is limited
to the number of GPUs in a node due to performance
considerations~\cite{megatronlm-2}. Nevertheless, our framework can still
support $4\times$, $6\times$ and $8\times$ larger base models on Perlmutter,
Summit and an NVIDIA-DGX-A100 machine respectively.

\section{Memory Savings via Tiling}
\label{sec:mem-opt}
In the previous section, we provided an overview of how TED distributes the
parameters and the computation of the forward and backward passes across the
GPUs. However, a naive combination of tensor, expert, and data parallelism
leads to significant spikes in memory usage during the optimizer step.
Interestingly, the magnitude of this spike becomes worse as we increase the
number of experts and/or the base model sizes. Note that it is important to
resolve this issue so that we are able to fit MoEs with large base models in
memory. Below, we discuss this phenomenon in detail and outline our solution to
resolve this issue.

\begin{figure}[h]
    \centering
      \includegraphics[width=\columnwidth]{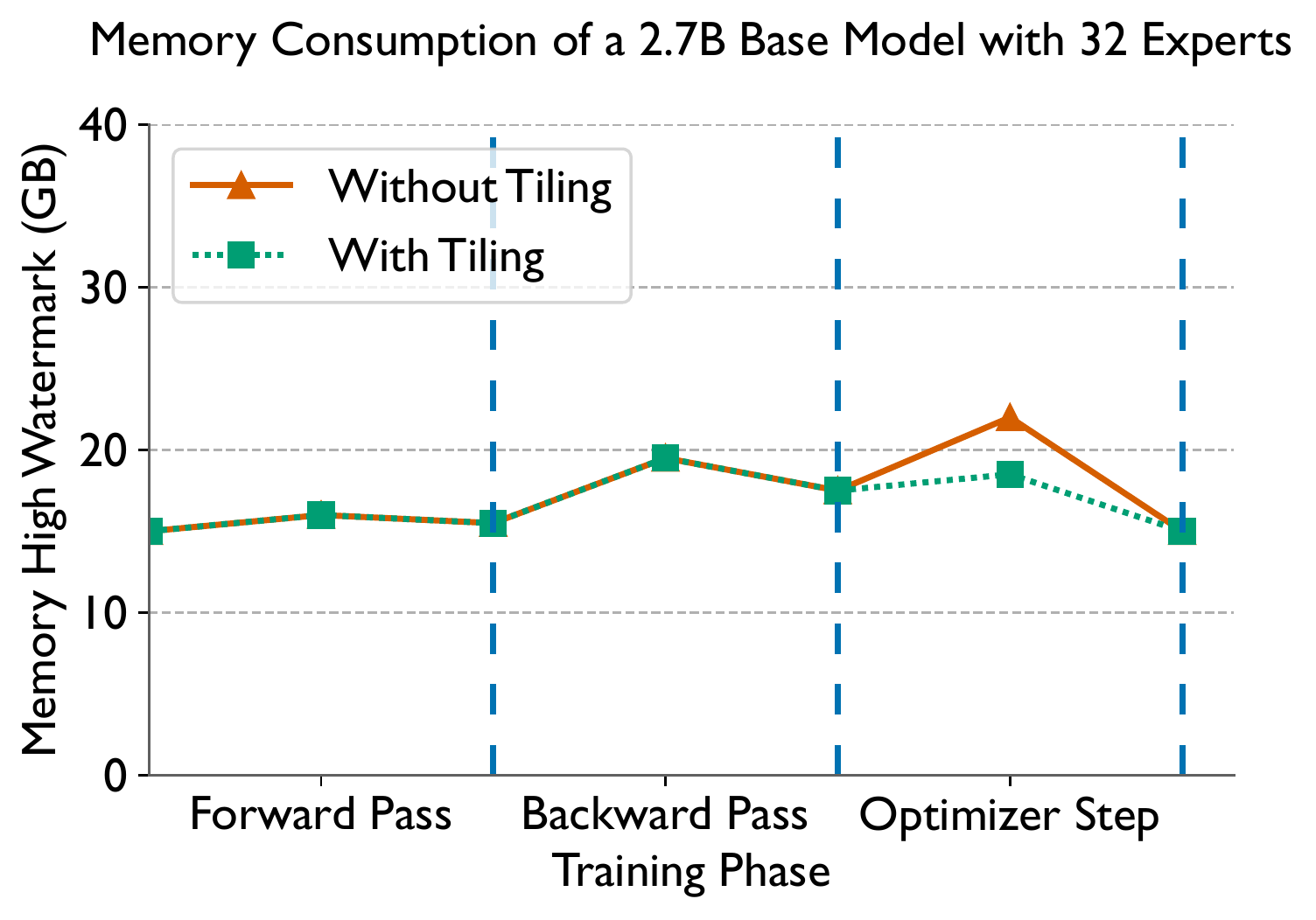}
    \caption{Memory consumption in the various phases of training for an MoE
with a 2.7B parameter base model and 32 experts on 32 GPUs of an NVIDIA
DGX-A100 (40 GB) cluster. We observe a large spike of an additional 4.5 GB in
memory usage during the optimizer step (red), which is significantly reduced to
around 1.5 GB by our tiled optimizer (green).}
\label{fig:spike}
\end{figure}

To demonstrate the aforementioned memory usage spike, we profile the memory
consumed per GPU during various phases of training (forward pass, backward
pass, optimizer step) for an MoE model with a 2.7B parameter base model and 32
experts, and show the results in Figure~\ref{fig:spike}. We run this experiment
on 32 GPUs of an NVIDIA DGX-A100 cluster with eight GPUs per node. We set the
degree of tensor and expert parallelism to 1 and 32 respectively. This results
in degrees of data parallelism as 32 and 1 for the non-expert and expert blocks
respectively.  We observe that memory consumption peaks during the optimizer
step with a very significant spike of around 4.5 GB. An intermediate step in
the optimizer phase in mixed precision training is the up-casting of 16-bit
gradients to 32-bit gradients before the optimizer updates the weights. This
requires the creation of a temporary buffer to store the 32-bit gradients and
is exactly the reason why there is a significant increase in memory
consumption. In fact, this problem becomes worse with increasing base model
sizes and/or expert counts.  Let us now understand why. 

TED uses ZeRO stage-1 which reduces memory consumption by sharding the
optimizer states and computation across the data parallel groups. Greater the
degree of data parallelism, the greater the reduction in memory
consumption~\cite{sc2020zero}.  From the discussion in
Section~\ref{sec:method}, we know that TED employs different degrees of data
parallelism for the expert parameters and non-expert blocks. In fact, it
follows from Equation~\ref{eqn:relation} that
\begin{align}
  \mathit{{G}_{tensor}}\times\mathit{G_{expert}}\times\mathit{G^{exp}_{data}}&=\mathit{{G}_{tensor}}\times\mathit{G^{nonexp}_{data}} \nonumber \\
  \mathit{G_{expert}}\times\mathit{G^{exp}_{data}}&=\mathit{G^{nonexp}_{data}} \nonumber \\
  \mathit{E}\times\mathit{G^{exp}_{data}}&=\mathit{G^{nonexp}_{data}} \nonumber
\end{align}

\begin{align}
  \mathit{G^{exp}_{data}}&=\frac{\mathit{G^{nonexp}_{data}}}{\mathit{E}} \label{eqn:dp}
\end{align}

From Equation~\ref{eqn:dp}, we can conclude that the degree of data parallelism
for the expert blocks is $\mathit{E}\times$ less than that for the non-expert
blocks. Therefore, ZeRO provides lesser memory savings for the expert blocks
than the non expert blocks. This is because the optimizer states for the expert
blocks are sharded over $\mathit{E}\times$ lesser GPUs.  Thus, as $\mathit{E}$
increases each GPU has to process increasing number of parameters in the
optimizer step. This leads to an increase in the size of the temporary 32-bit
gradient buffer required to up-cast the expert parameter gradients. Increasing
the base model size also worsens this problem as the size of the expert
parameter group is directly proportional to the base model size. This is why it
is imperative to resolve this issue such that we can train MoEs with large base
models and/or large number of experts.

In this work, we propose a tiled formulation of the optimizer that strives to
alleviate the aforementioned issue. Instead of processing the entire expert
parameter group at once, we propose partitioning these parameters into
``tiles'' of a predefined size and iteratively processing these tiles. This
ensures that at any given time, temporary 32-bit gradients are only produced
for parameters belonging to a given tile.  The temporary memory used to store
these gradients can in fact be reused across tiles. For a tile size
$\mathit{ts}$, we now only need $4\times \mathit{ts}$ bytes of memory to
materialize the 32-bit gradients. This makes the optimizer memory spike
independent of the number of experts and the base model sizes! In our
experiments, we fix the tile size to 1.8 million parameters, which essentially
caps the spike in the optimizer step to 1 GB. We observed that this tile size
is large enough to not cause any performance degradation due to the latency of
multiple kernel launches. In Figure~\ref{fig:spike}, we demonstrate how our
tiled optimizer reduces the per GPU peak memory consumption for the
aforementioned MoE with a 2.7B parameter base model and 32 experts by 3 GB. In
fact, on another MoE with 6.7B parameters and 16 experts on 32 GPUs, our
framework ran out of memory without tiling. Whereas, with tiling enabled, we
were able to successfully train this model with a peak memory consumption of
31.3 GB. Since the maximum memory capacity of these GPUs is 40 GB, optimizer
tiling provides a significant memory savings of more than 21.75\%!

\section{Performance Optimizations}
\label{sec:perf-opt}
\begin{figure}[h]
    \centering
      \includegraphics[width=\columnwidth]{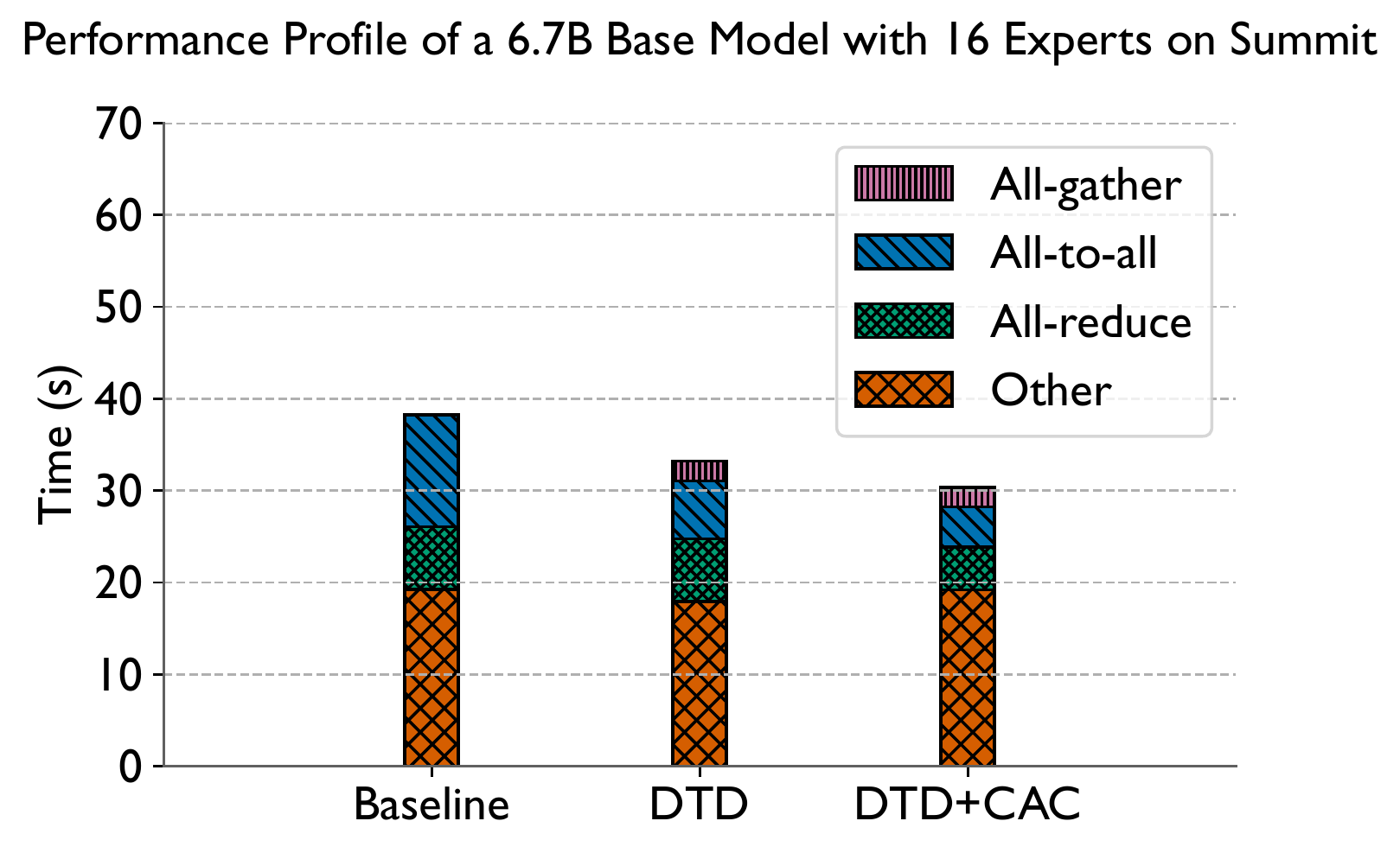}
    \caption{Impact of our communication optimizations on the batch time of an
MoE model with a 6.7B parameter base model and 32 experts on 128 GPUs of Summit
(batch size: 1024). Our optimizations result in significant
reductions of 64.12\% and 33\% in the all-to-all and all-reduce time
respectively, thereby improving the overall training time by 20.7\%.}
\label{fig:breakdown}
\end{figure}

In the preceding sections, we focused on increasing the maximum possible size
of MoEs that are supported by our framework.  While the memory savings provided
by expert and tensor parallelism contribute to this, they also result in a
significant portion of the batch time being spent in expensive collective
communication. In Figure \ref{fig:tp-ep-dp}, we can observe that the forward
pass includes two all-reduce calls within the tensor parallel groups, and two
all-to-all calls within the expert parallel groups. During the backward pass,
these calls are repeated again. Also, large model training almost always uses
activation checkpointing~\cite{chen2016training}, which significantly reduces
activation memory at the expense of a duplicate forward pass per layer. Thus,
overall we end up with six all-to-all and six all-reduce communication calls,
which become a significant bottleneck in training. We empirically demonstrate
this in Figure \ref{fig:breakdown} (leftmost bar titled Baseline), wherein we
observe that almost half of the batch time is spent in the all-to-all and
all-reduce communication calls.  We will now describe two performance
optimizations that seek to reduce the time spent in these communications and
are extremely critical to the performance of our framework.  

\subsection{Duplicate Token Dropping (DTD) for Reducing Communication Volume}
\label{sec:dtd}

MegatronLM's tensor parallelism for partitioning self-attention and feed
forward blocks involves issuing an all-reduce on local partial outputs to
materialize the full outputs on each rank~\cite{megatronlm}. For example, in
Figure \ref{fig:tp-ep-dp}, GPUs 0 and 1 issue an all-reduce (\circled{2}) after
the self-attention block to assemble the full self-attention outputs for tokens
1 and 2. While, this leads to duplication of activations across the tensor
parallel ranks, it is not an issue for training regular transformer models
(i.e. without experts) as the tensor parallel blocks under MegatronLM's
algorithm require a complete set of input activations on each tensor parallel
rank. Thus the   duplicate activations output by a tensor parallel block serve
as the required input for its successor. However, for MoEs, an unwanted side
effect of this design choice is the presence of redundant tokens in the
all-to-all communication calls.  For example consider the first all-to-all in
Figure \ref{fig:tp-ep-dp} (\circled{4}). Self-attention output activations, a1
and a2, are communicated by both GPUs 0 and 1. Similarly, GPUs 2 and 3 both
communicate a3 and a4.  In general, the amount of unnecessary data in the
all-to-all communication calls for a given token is proportional to the degree
of tensor parallelism. Thus, naively combining expert and tensor parallelism
can lead to the all-to-all communication becoming a significant bottleneck,
especially as we try to increase the base model sizes (larger base models need
more tensor parallelism). For example, in Figure \ref{fig:breakdown}, 32\% of
the batch time is spent in the all-to-all (leftmost bar titled baseline)! The
degree of tensor parallelism and thus the degree of redundancy in the
all-to-alls is four here. 

\begin{figure}[h]
  \centering
    \includegraphics[width=\columnwidth]{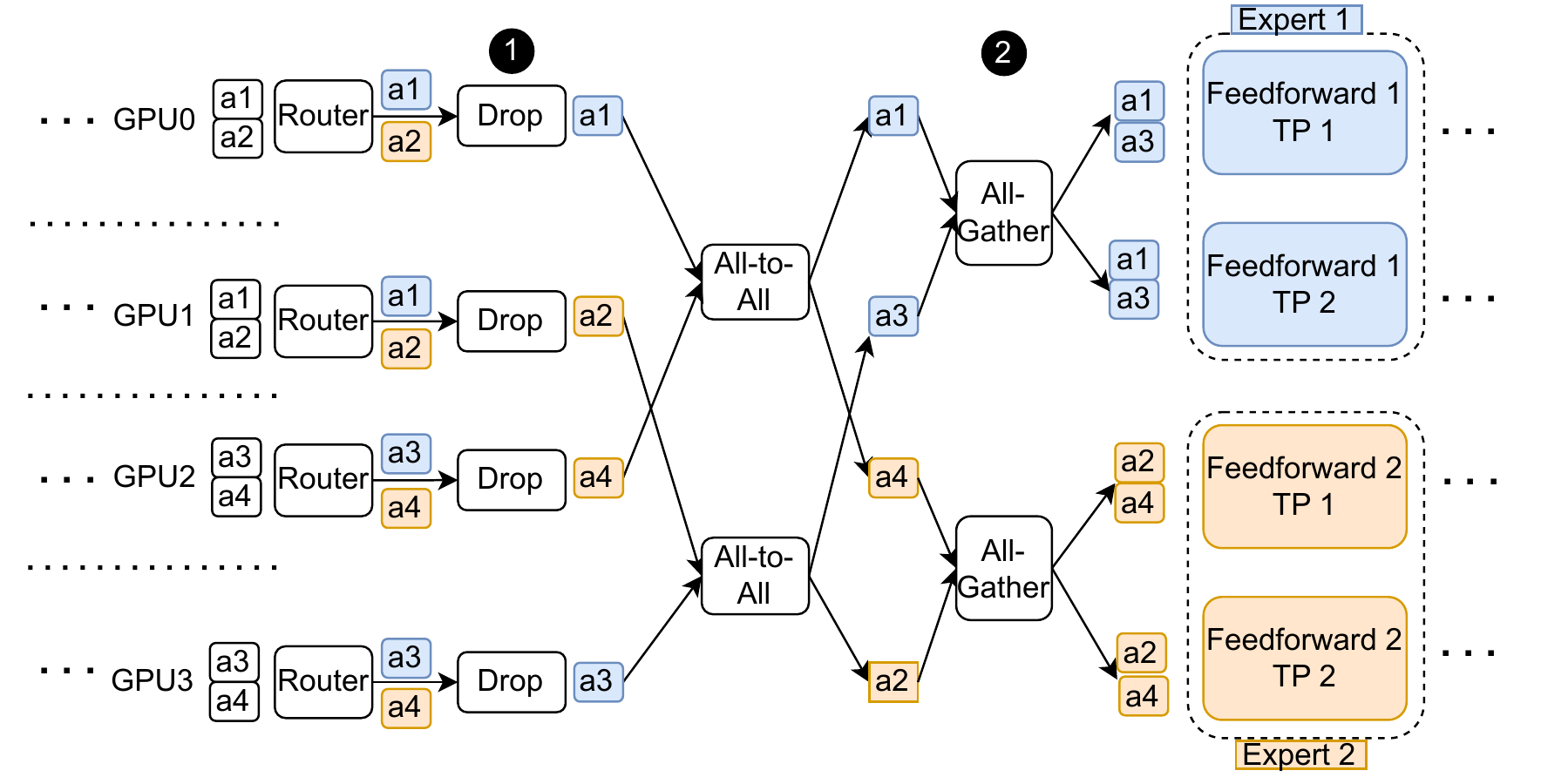}
  \caption{Duplicate token dropping (DTD) in the first all-to-all communication
of an MoE layer (Steps 3--5 in Figure \ref{fig:tp-ep-dp}). Before the
all-to-all, we apply the drop operation, which eliminates redundant tokens
across tensor parallel ranks, and reduces the all-to-all message sizes by the
degree of tensor parallelism. After the all-to-all, GPUs reassemble the full
input to the feed forward blocks by issuing an all-gather between the tensor
parallel ranks.}
\label{fig:dtd}
\end{figure}

To resolve this bottleneck, we propose duplicate token dropping (DTD), a
communication optimization geared towards eliminating unnecessary data in the
all-to-all communication. We illustrate the working of DTD in Figure
\ref{fig:dtd} for the first all-to-all communication in an MoE layer
(\circled{4} in Figure \ref{fig:tp-ep-dp}). Before the all-to-all is issued,
GPUs within tensor parallel groups participate in a ``drop'' operation
(\circled{1} in Figure \ref{fig:dtd}). The drop operation ensures that the
there is no redundancy in the output activations across the tensor parallel
ranks. For instance, GPU 0 drops the activation of a2 whereas GPU 1 drops the
activation a1, thereby completely eliminating redundancy within their tensor
parallel group. Similarly, GPUs 3 and 4 drop a3 and a4 respectively. The drop
operation thus reduces the all-to-all message sizes by two times in this
example, and in general the reduction is equal to the degree of tensor
parallelism. However, after the all-to-all, the GPUs do not have the full input
activations to commence the computation of the expert feed forward blocks. For
instance, GPU 0 has the input activations for the token 1, but not for token 3
and vice versa for GPU 1. Therefore, to assemble the full input activations, we
issue an all-gather call (\circled{2} in Figure \ref{fig:dtd}) between the
tensor parallel GPUs. The all-gather ensures that the input dependencies for
the expert feedforward blocks are met. 

During the backward pass the all-gather call is replaced by a drop operation
and the drop operation is replaced by an all-gather call.  For the MoE model in
Figure \ref{fig:breakdown}, we observe that DTD reduces the all-to-all
communication time by 48\%. While the inclusion of DTD leads to an additional
all-gather operation (shown in red on top of the second bar), this overhead is
outweighed by the improvement in the all-to-all communication timing. Overall,
DTD results in an improvement of 13.21\% in the batch time.

\subsection{Communication-aware Activation Checkpointing (CAC)}

We now turn our attention to a second source of redundant communication in
large model training, namely activation checkpointing \cite{chen2016training}.
Intermediate activations in a neural network generated during the forward pass
need to be stashed in memory as they are required during the backward pass for
the gradient computation. However, for large model training, storing all the
activations can lead to tremendous memory overhead. Activation checkpointing
alleviates this issue by storing only a subset of the activations, which are
essentially just the input activations of every layer. During the backward pass
of a layer, the remaining activations are re-materialized from its stashed
input activation by doing a local forward pass for that layer. Thus, activation
checkpointing saves activation memory at the expense of a duplicate forward
pass for every layer, and is almost always used for training large neural
networks.  For more details, we refer the reader to Chen et
al.~\cite{chen2016training}.

We know from Section \ref{sec:method} that the forward pass of an MoE layer in
TED involves two all-to-alls and two all-reduce calls in the forward pass and
two all-to-alls and two all-reduce calls in the backward pass. Since activation
checkpointing involves repeating the forward pass of a layer, we now end up
with two additional all-to-all and all-reduce calls, thereby increasing
communication volume by $1.5\times$ and making the training process
inefficient. 

To this end, we propose communication-aware checkpointing (CAC), a
communication optimization that eliminates the additional communication in the
second forward pass induced by activation checkpointing. During the first
forward pass, CAC stashes the outputs of each all-reduce and all-to-all
communication call along with the data stashed by standard activation
checkpointing. Now, during the second forward pass, we bypass these
communication calls and instead return the outputs for these communication
calls stashed during the first forward pass. CAC thus reduces the communication
volume by 33\% at the expense of using extra GPU memory. For the MoE model in
Figure \ref{fig:breakdown}, CAC indeed reduces the all-to-all and all-reduce
communication times by 33\% (compare second and third bars). In combination
with DTD, the reductions in the all-to-all and all-reduce communication times
are 64.12\% and 33\% respectively, amounting to a speedup of nearly 20.7\% over
the baseline version of DeepSpeed-TED.

\section{Experimental Setup}
\label{sec:setup}
This section provides an overview of the empirical evaluation of DeepSpeed-TED.
Our framework is open source, and has been integrated in DeepSpeed, a
state-of-the-art framework for parallel deep learning. We conduct our
experiments on the Summit and ThetaGPU supercomputers.  Summit has six 16 GB
NVIDIA V100 GPUs per node, each having a peak half precision throughput of 125
Tflop/s. Each node has two 22-core Power 9 CPUs.  The peak intra-node and
inter-node GPU bidirectional communication bandwidths are 50 GB/s (NVlink) and
25 GB/s (Infiniband) respectively. On the other hand, ThetaGPU is a NVIDIA DGX
A100 machine with eight 40 GB NVIDIA A100 GPUs per node, each having a peak
half precision throughput of 312 Tflop/s. On this machine, the peak intra-node
and inter-node GPU bidirectional communication bandwidths are 600 GB/s (NVlink)
and 200 GB/s (Infiniband) respectively.

\subsection{Neural Network Architectures and Datasets}

Table~\ref{tab:setup-perf} lists the various base model architectures used in
this study. All MoEs used in our empirical experiments are constructed by
adding expert blocks to every alternate layer of one of these base models (this
is in line with previous work~\cite{g-shard, switch-transformer, ds-moe-ml}).
The base models and their corresponding hyperparameters are taken from Brown et
al.~\cite{gpt-3}. We use the Pile dataset to generate input
tokens~\cite{the_pile}. We use the AdamW optimizer~\cite{adamw}, which is the
standard practice for large language model training. We implement the layers of
our transformer models using MegatronLM's GPU kernels~\cite{megatronlm}.

\begin{table}[h]
    \centering 
    \caption{Architectural details and batch sizes for the various transformer base models used to build the MoEs used in this study. All hyperparameters including the 
    batch sizes are taken from Brown et al.~\cite{gpt-3}.\label{tab:setup-perf}}
    \begin{tabular}{rcccr}
    \toprule
    \# Parameters & \# Layers & \multicolumn{1}{c}{\begin{tabular}[c]{@{}c@{}}Hidden \\ Size\end{tabular}} & \multicolumn{1}{c}{\begin{tabular}[c]{@{}c@{}}Attention \\ Heads\end{tabular}} & Batch Size \\ \midrule
    1.3B       & 24                                                                              & 2048        & 16                                                                             & 512 \\
    2.7B       & 32                                                                              & 2560        & 32                                                                             & 512 \\
    6.7B       & 32                                                                              & 4096        & 32                                                                             & 1024 \\
    13.0B        & 40                                                                              & 5140        & 40                                                                             & 2048 \\
    \bottomrule    
    \end{tabular}
\end{table}

First, we establish the correctness of our implementation by training a 2.6B
parameter MoE model (1.3B parameter base model and 4 experts) to completion on
the BookCorpus dataset~\cite{book-corpus}, on 8 GPUs, and present the
validation loss curves.  For reference, we also train this model using
DeepSpeed-MoE~\cite{ds-moe-systems}, the current state-of-the-art framework for
training MoEs and compare the two loss curves. Then, we demonstrate the maximum
MoE model sizes that our framework can support for a given number of GPUs and
compare it with DeepSpeed-MoE~\cite{ds-moe-systems}. Next, we conduct strong
scaling studies using MoEs built from the 1.3B, 2.7B and 6.7B parameter
transformer models in Table \ref{tab:setup-perf} on 32 to 256 GPUs. At 32 GPUs,
we add as many experts as the system memory permits.  We strong scale a model
in two ways, first increasing the number of GPUs while keeping the number of
experts constant, and second by varying the number of experts proportional to
the number of GPUs. Note that even though the latter experiment increases the
model size with scale, it is still considered strong scaling as adding experts
to a base model does not change the total number of floating point operations
in training. For our weak scaling runs, we fix the number of experts to 16 and
use base models of increasing sizes from Table \ref{tab:setup-perf} as we go
from 32 to 256 GPUs. Note that this is weak scaling, because the number of
floating point operations are proportional to the base model size.

\subsection{Evaluation Metrics}

We illustrate the results of our experiment using the average time per
iteration (or batch).  To calculate this, we first run a given model with 100
batches sampled from the Pile dataset \cite{the_pile} and take the average of
the last 90. We do not include the first 10 batches because PyTorch issues
expensive mem-alloc calls to the CUDA runtime in the initial iterations to
reserve enough memory for training. We also derive the percentage of peak
half-precision throughput from the average batch time using Narayanan et al.'s
formulation\cite{megatronlm}. Note that this an analytical formulation of the
total number of flop/s (and thus the percentage of peak half-precision
throughput).  Since Narayanan et al.'s formulation is a lower bound on the
total floating point operations, we expect empirically measured flop/s to be
higher.

\section{Results}
In this section, we discuss the results of the empirical experiments outlined in Section \ref{sec:setup}.

\subsection{Validating Our Implementation}

\begin{figure}[h]
  \centering
    \includegraphics[width=\columnwidth]{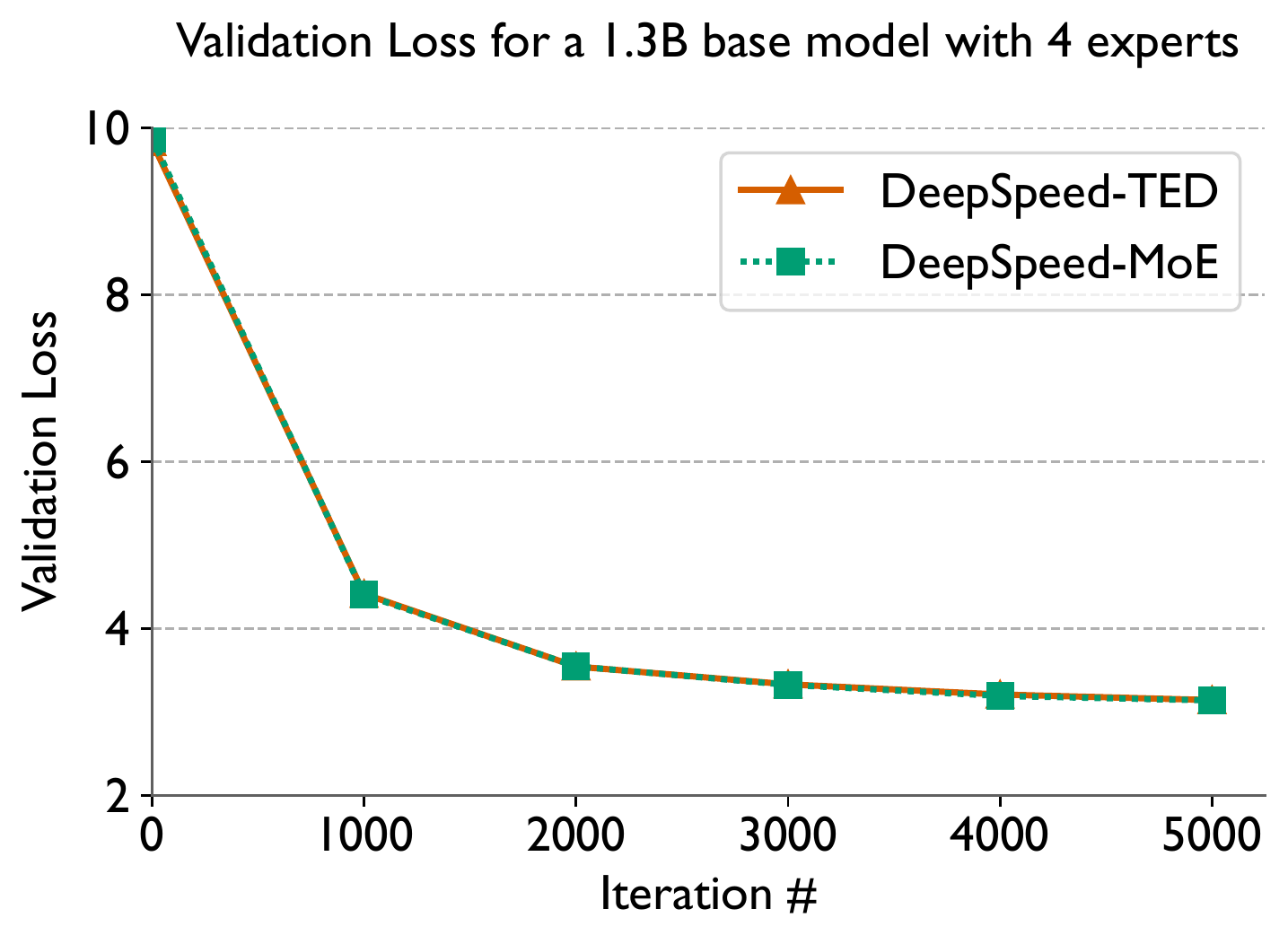}
  \caption{Validation loss for an MoE with a 1.3B base model and four experts on eight GPUs of ThetaGPU on the BookCorpus 
  dataset~\cite{book-corpus}. We use a batch size of 128 and sequence length of 2048. We set 
  $\mathit{G_{tensor}}=2$, $\mathit{G_{expert}}=4$, $\mathit{G^{exp}_{data}}=1$, 
  $\mathit{G^{nonexp}_{data}}=4$.}
\label{fig:stat-eff}
\end{figure}

To verify the correctness of DeepSpeed-TED, we train an MoE with a 1.3B parameter base model and 4 experts on 
8 GPUs of ThetaGPU and present the validation loss curve in Figure~\ref{fig:stat-eff}.
We set $\mathit{G_{tensor}}=2$, $\mathit{_{expert}}=4$, $\mathit{G^{nonexp}_{data}}=4$, and $\mathit{G^{exp}_{data}}=1$.  
This allows us to test the correctness of our framework in a scenario where all three dimensions of its hybrid parallel 
approach are active. We also enable the communication optimizations discussed in Section~\ref{sec:perf-opt} i.e. DTD and CAC. 
We observe that our framework is able to successfully train the model to 
convergence, and produces identical loss curves to DeepSpeed-MoE, a system that has been previously used to train 
state-of-the-art MoE models. In this way, we establish the correctness of our implementation. 

\subsection{Comparison of Supported Model Sizes}

\begin{figure*}[t]
  \centering
  \includegraphics[width=0.33\textwidth]{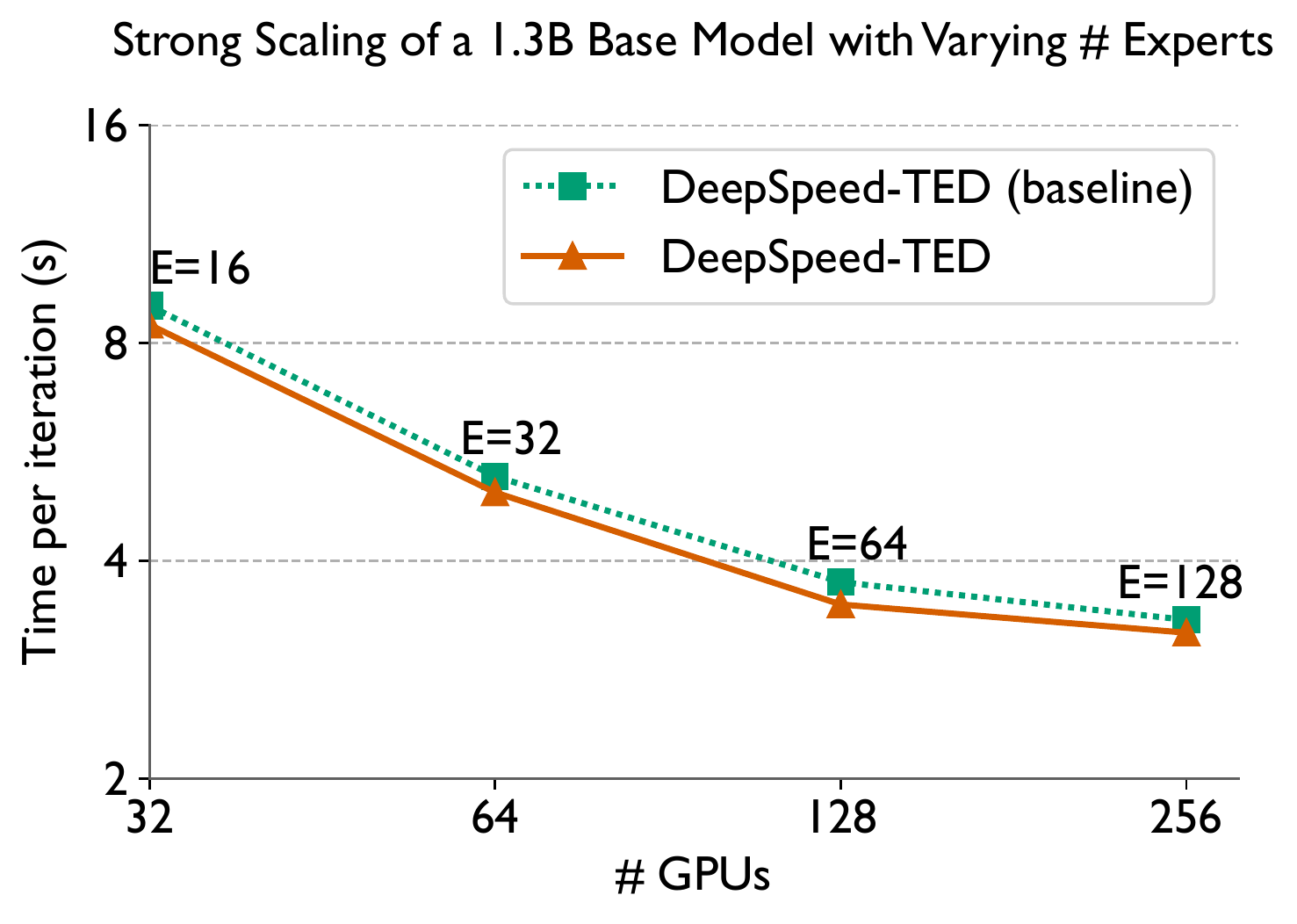}
  \includegraphics[width=0.33\textwidth]{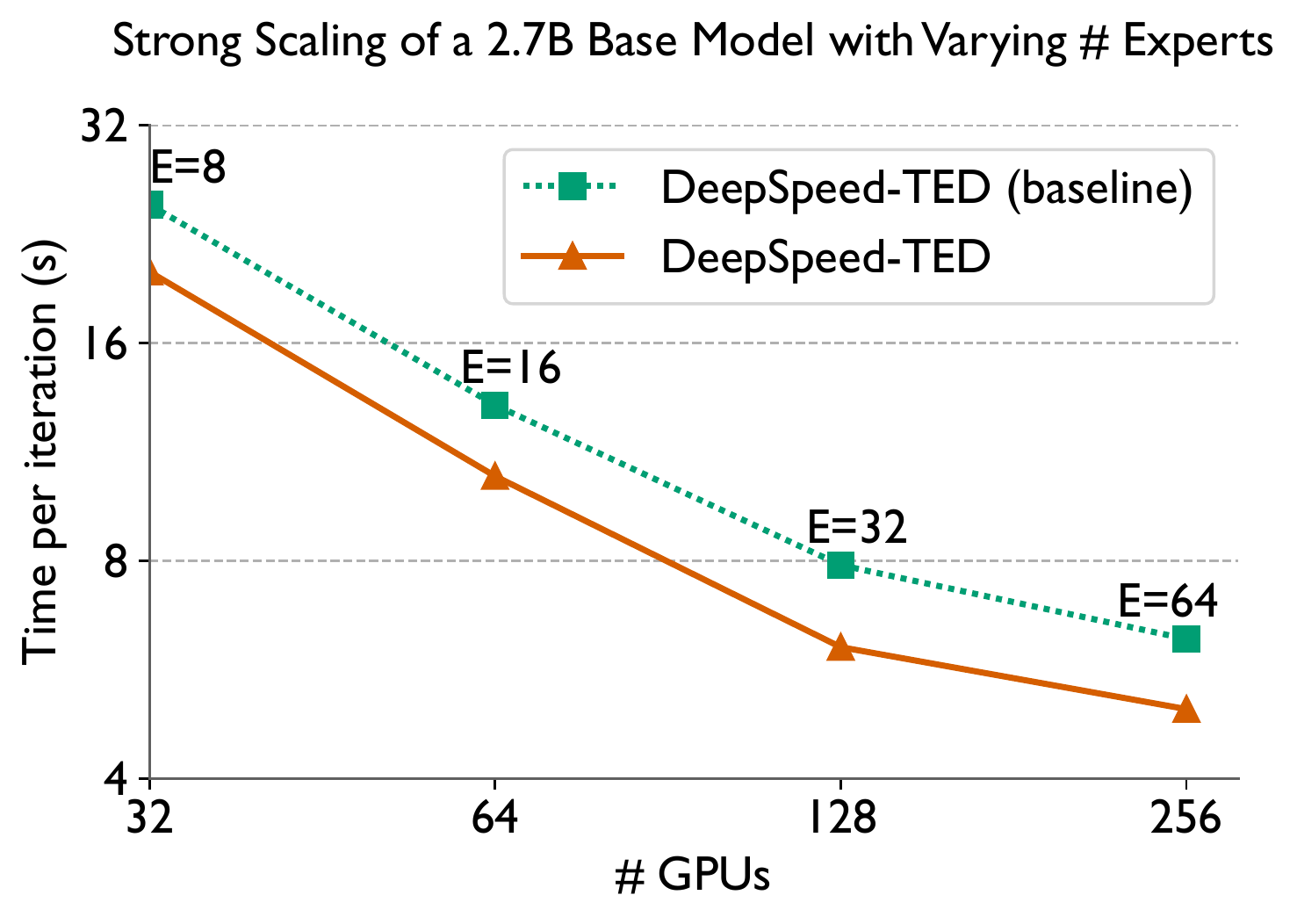}  
  \includegraphics[width=0.33\textwidth]{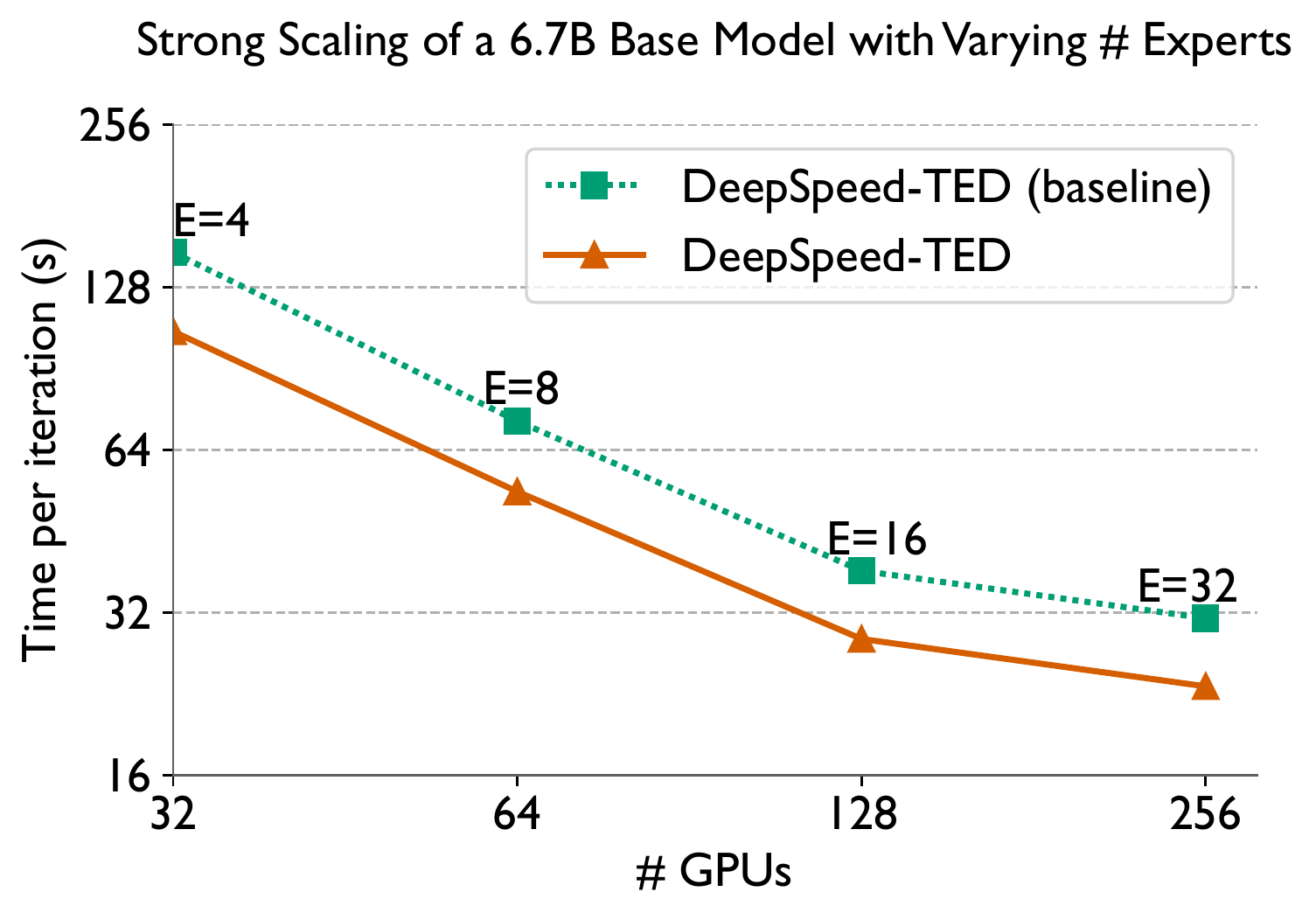}
  \caption{Strong scaling (with varying number of experts) of MoEs with the 1.3B,
2.7B and 6.7B parameter models in Table \ref{tab:setup-perf} used as base, on
V100 GPUs of Summit.  We annotate the plot with the number of experts used at
each GPU count. We sample input batches of sizes 512, 512 and 1024 respectively,
from the Pile dataset\cite{the_pile}.}
\label{fig:scale-2.7B-6.7B}
\end{figure*} 

\begin{figure}[h]
    \centering
      \includegraphics[width=\columnwidth]{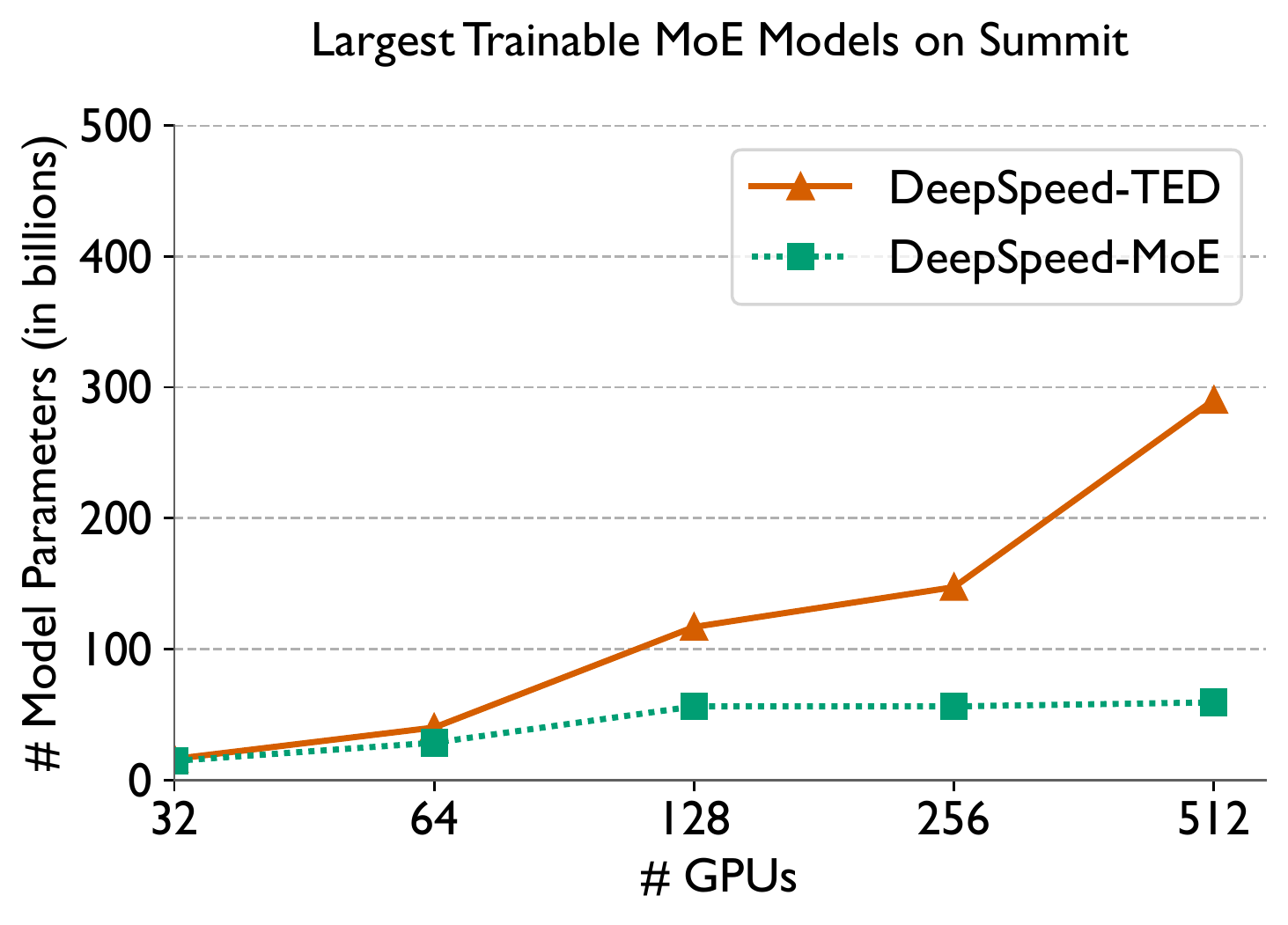}
    \caption{Largest MoE model sizes supported on various GPU counts on Summit.
We construct MoEs using base models from Table \ref{tab:setup-perf} and number
of experts in the range of 4 to 128. Compared to DeepSpeed-MoE
\cite{ds-moe-systems}, our framework supports 1.09-4.8$\times$ larger MoE
models, with the ratio increasing with increasing number of GPUs.}
\label{fig:scale-mem}
\end{figure} 

Figure \ref{fig:scale-mem} illustrates the results of our experiment in which we benchmark the largest MoE models that our framework and DeepSpeed-MoE~\cite{ds-moe-systems} 
can train without running out of memory for various GPU counts ranging between 32 and 512. We use the base models in Table \ref{tab:setup-perf}. 
To make sure that the experiment is fair to both the frameworks we do two things. While our proposed approach can in theory support arbitrarily large base models 
by increasing the degree of tensor parallelism, it is well known that tensor parallelism is extremely inefficient when used across nodes. Therefore, we only allow 
our framework to use a maximum tensor parallel degree of six, which is the number of GPUs on a node of Summit. Second, we limit the largest 
possible number of experts to 128 as prior work has demonstrated limited improvements in the statistical efficiency of a model beyond this number 
\cite{switch-transformer}. 

Across the range of GPUs used in this experiment, we observe that DeepSpeed-TED supports $1.09-4.8\times$ larger MoE models than DeepSpeed-MoE. 
We also observe that this ratio increases as we increase the number of GPUs. This can be explained by Equation \ref{eqn:lower-bound-full}, which states that the 
memory consumption of our approach decreases with increasing number of GPUs. We observe that beyond 128 GPUs, our proposed framework can train  
MoEs with hundreds of billion of parameters on Summit, which is not possible with DeepSpeed-MoE. Thus, we have empirically demonstrated how our DeepSpeed-TED
can enable the development of high quality MoE models, the parameters of which have been scaled along the base model dimension as well as the expert dimension. 

\subsection{Strong Scaling Performance}

We now discuss the results of our strong scaling experiments, starting with the runs that varied the number of experts proportional to 
the number of GPUs. We demonstrate the results for the 1.3B, 2.7B and 6.7B base models in
Figure \ref{fig:scale-2.7B-6.7B}. To demonstrate the efficacy of the communication optimizations discussed in Section
\ref{sec:perf-opt}, we also benchmark the baseline version of our framework i.e. with DTD+CAC disabled, and call it DeepSpeed-TED (baseline). Across all the figures, we observe 
that augmenting the training procedure with DTD and CAC indeed improves the hardware efficiency of training. However, while the speedups for the 2.7B and 6.7B parameter 
base models are significant: 19 to 23\% and 25 to 29\% respectively, our communication optimizations seem to be less effective for the smallest 1.3B 
base model providing modest speedups of around 4 to 7\%. This is because at the given GPU counts and number of experts, ZeRO's memory optimizations and 
expert parallelism are able to fit this model in memory without the aid of tensor parallelism. Without tensor parallelism there is no redundancy in the all-to-all 
communication (see Section \ref{sec:dtd}) and thus the DTD communication optimization is of no use in this scenario. Similarly, without tensor parallelism there is no 
all-reduce communication (\circled{2} and \circled{6} of Figure \ref{fig:tp-ep-dp}). Thus, CAC only eliminates the unnecessary all-to-all calls, and is only 
partially applicable to this scenario. This explains the reduced effectiveness of our optimizations for the 1.3B base model.

Unlike the 1.3B base model, 
the 2.7B and 6.7B model require a tensor parallel degree of 2 and 4 to fit in available GPU memory. The ensuing redundancy in the all-to-all messages and the introduction 
of tensor parallelism thus makes our communication optimizations quite effective. Again, we observe larger speedups for the MoEs using the 6.7B base model (25--29\% versus 19--23\%)
as a higher degree of tensor parallelism implies more redundancy in the all-to-all messages, which our optimizations successfully eliminate. It also implies a larger proportion of 
time spent in tensor parallel all-reduces which is significantly reduced by CAC. 

\begin{figure}[h]
  \includegraphics[width=\columnwidth]{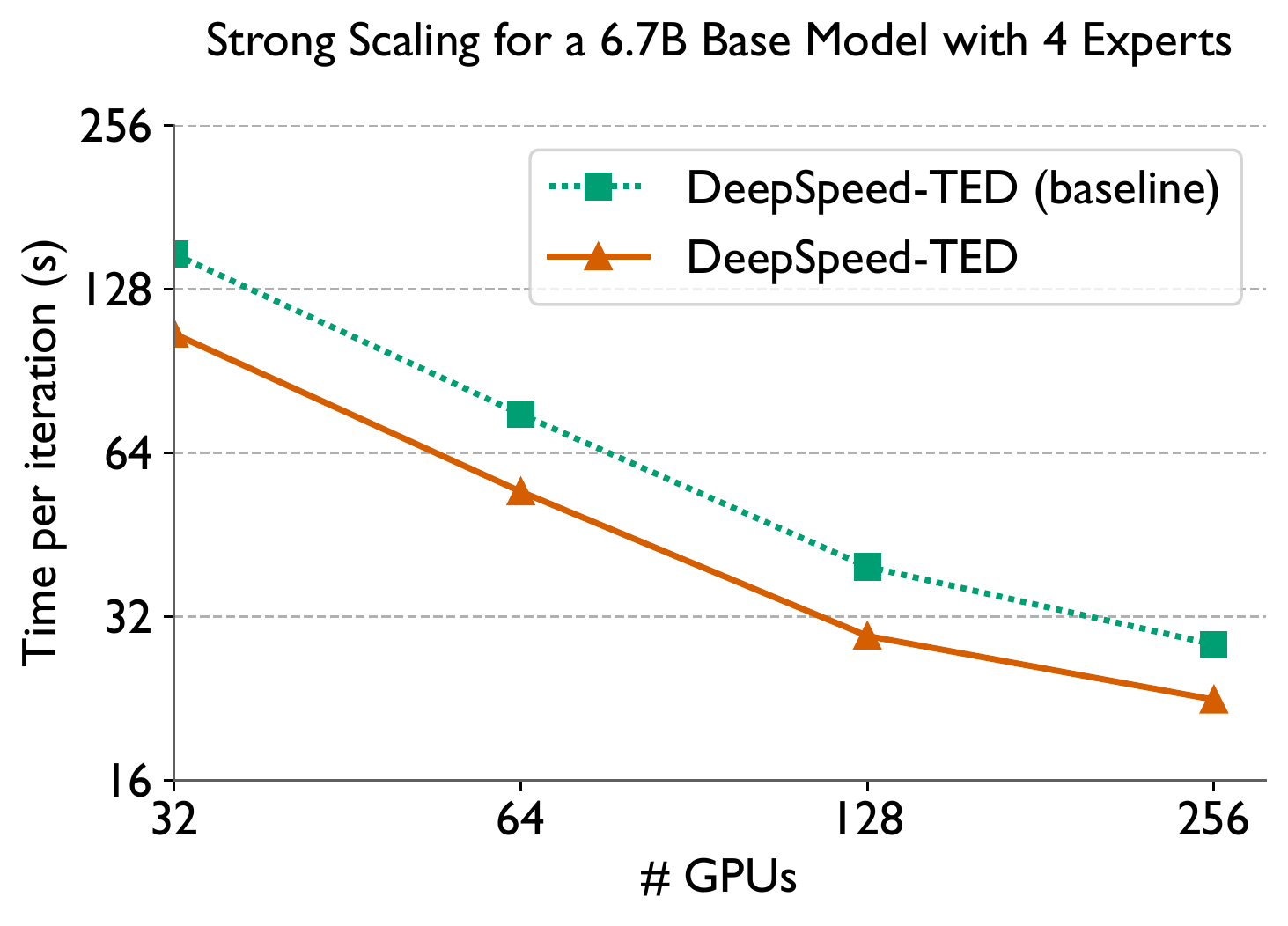}
  \caption{ Strong scaling (with number of experts fixed to four) of a MoE with a 6.7B parameter model in Table \ref{tab:setup-perf} used as base, on V100 GPUs of Summit. 
  We sample input batches of size 1024 from the Pile dataset\cite{the_pile}.
  \label{fig:scale-6.7B-fixed}}
\end{figure}

In our strong scaling runs with fixed number of experts, we observed very similar absolute times per iteration and relative speedups for all the three models. For brevity, we only 
include the results for the 6.7B parameter base model, and illustrate them in Figure \ref{fig:scale-6.7B-fixed}. We have thus verified that our optimizations are effective at 
improving performance in two strong scaling setups across various base model sizes. 

\subsection{Weak Scaling Performance}

\begin{figure}[h]
  \includegraphics[width=\columnwidth]{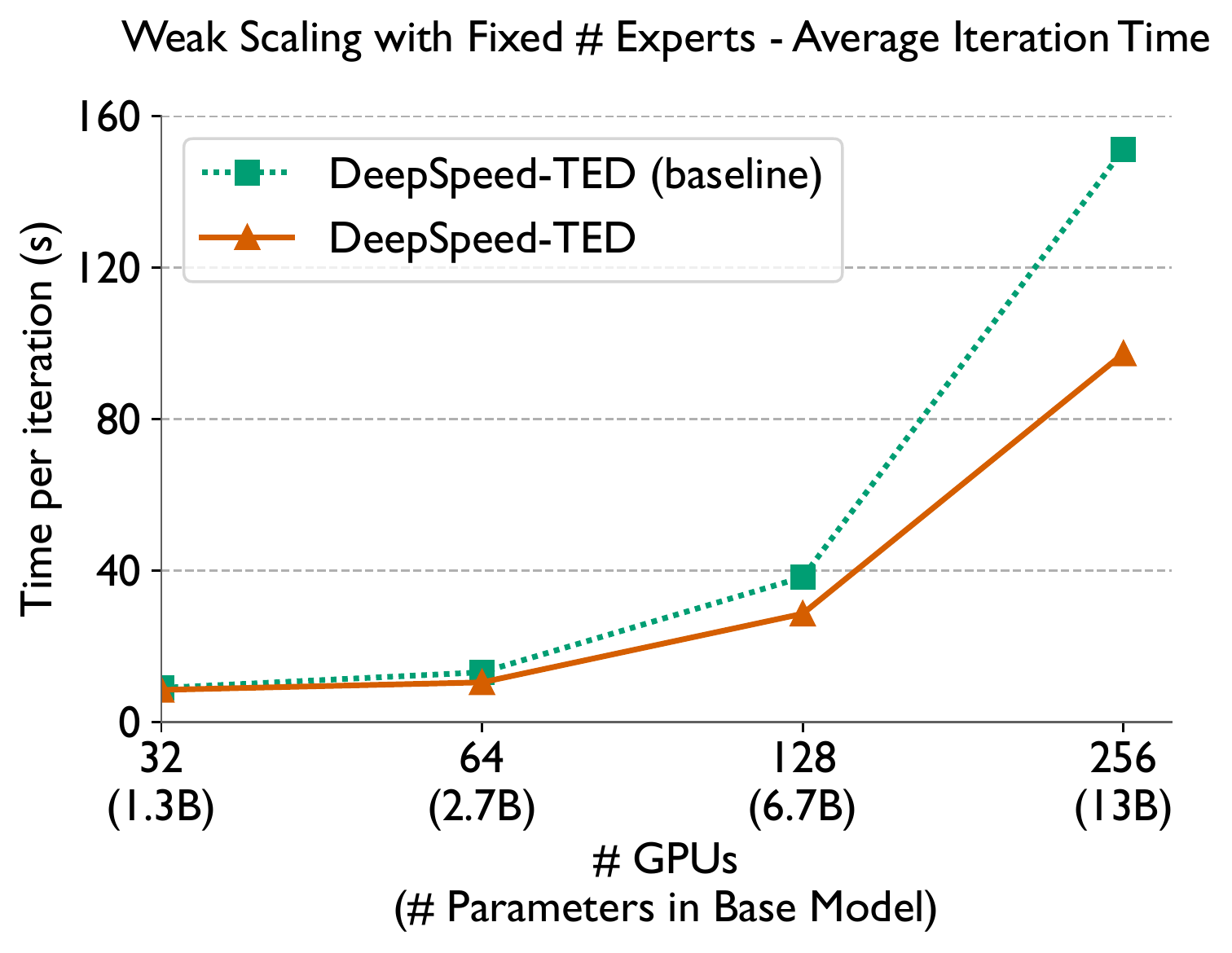}
  \caption{ Average time per iteration (left) for a weak scaling 
  study of MoE models with 16 experts on Summit. Base models and batch sizes are taken from Table \ref{tab:setup-perf}.
  \label{fig:weak-scale}}
\end{figure}

As discussed in Section \ref{sec:setup}, we conduct a weak scaling experiment by fixing the number of experts to 16 and varying 
the base model size in proportion with the number of GPUs. We demonstrate the time per iteration (or batch) and percentage of peak half-precision throughputs for this experiment in Figure \ref{fig:weak-scale} and Table~\ref{tab:weak-scale} respectively. Again, 
we observe a minor speedup of 6\% for the 1.3B base model, and significant speedups of 20\%, 25\% and 36\% for the 2.7B, 
6.7B and 13B base models respectively. Just like the previous section, the progressively increasing effectiveness of our 
communication optimizations for larger base models can be explained by the correspondingly increasing degrees of tensor parallelism - 1, 2, 4, and 8. This creates more redundancy for the larger models in the all-to-all and increases the net communication volume of 
the all-reduces. Note that while the speedups for the 13B parameter model is significant (36\%), the hardware utilization for this model 
is extremely low. Even with our optimizations, we are only able to achieve 11.7\% of the peak half-precision flop/s, which is 
significantly lower than the 1.3B (37\% of peak), 2.7B (30\% of peak) and 6.7B (27\% of peak) base models. The explanation for this 
observation is that a tensor parallel degree of 8 for this model is greater than the number of GPUs on a Summit node.
This experiment corroborates prior work which has observed that Megatron-LM's algorithm does not scale well beyond the 
confines of a node \cite{megatronlm-2, singh:ipdps2022}. 

\begin{table}[h]
  \centering 
  \caption{Percentage of peak half precision throughput for a weak scaling 
  study of MoE models with 16 experts on Summit. Base models and batch sizes are taken from Table \ref{tab:setup-perf}. \label{tab:weak-scale}}
  \begin{tabular}{rrc}
  \toprule
          & Base Model Size & Throughput \\
  \# GPUs & (\# Parameters) & (\% of peak)  \\ \midrule
  32      & 1.3B         & 36.7      \\
  64      & 2.7B         & 30.0      \\
  128     & 6.7B         & 26.2      \\
  256     & 13.0B          & 11.7      \\ 
  \bottomrule
  \end{tabular}
\end{table}

\section{Related Work}
Due to the increasing computational costs of training state-of-the-art neural
networks, several frameworks and algorithms have been proposed that can train
these networks in parallel on networked multi-GPU clusters.  These can broadly
be divided intro three categories - data, tensor, and pipeline parallelism.
Under data parallelism, participating GPUs are assigned a full copy of the
neural network. The parallelism comes from the fact that each GPU works on a
equal sized shard of the input batch at every iteration. Perhaps due to its
simplicity of implementation, data parallelism has been the most widely adopted
algorithm and can be found in popular deep learning frameworks like PyTorch (as
Distributed Data Parallel~\cite{pytorchdist-vldb}). However, a major limitation
of data parallelism is that it requires the full neural network to fit on each
GPU. To resolve this issue, Rajbhandari et al.~propose the Zero Redundancy
Optimizer (ZeRO) which shards the parameters, gradients, and/or optimizer
states of the model across participating GPUs~\cite{sc2020zero}, and enable
training of much larger models that far exceed the memory capacity of a single
GPU. PyTorch also natively offers Fully Sharded Data Parallelism (FSDP), which
is based off a similar idea~\cite{pytorchFSDP}.

Tensor parallel algorithms like MegatronLM~\cite{megatronlm} divide the
parameters and computation of each layer of a neural network across
participating GPUs and can thus also be used to train neural networks that do
not fit on a single GPU. Other examples of tensor parallel frameworks and
algorithms are ~\cite{you-2d, you-2.5d, you-3d} for fully connected layers,
~\cite{spatial-parallel-cnn-dryden, channel-filter-parallel-cnn-dryden} for
convolution layers, and ~\cite{yellick-graph} for graph neural networks. On the
other hand, pipeline parallelism involves assigning the parameters and
computation of a contiguous subset of layers to each
GPU~\cite{huang2019gpipe_nips,megatronlm-2,zero_3D,rannc,singh:ipdps2022}.
Parallelism is achieved by breaking a batch into microbatches and processing
the microbatches in a pipelined fashion (akin to pipelining in computer
architecture). Narayanan et al.~show how combining tensor, pipeline, and data
parallelism can be an extremely efficient strategy to train large multi-billion
parameter models at scale~\cite{megatronlm-2}.  As future work, we plan to
integrate pipeline parallelism in DeepSpeed-TED to further enhance its
performance. 

To combat the rising computation costs of training state-of-the-art neural
networks like Chinchilla~\cite{chinchilla}, GPT-3~\cite{gpt-3}, and
Megatron-Turing NLG~\cite{megatron-turing-nlg-530b}, the machine learning
community has recently turned its attention to the Mixture-of-Experts (MoE)
architecture to train large compute efficient transformer models for natural
language processing and computer
vision~\cite{switch-transformer,glam-moe,vision-moe,g-shard}. Subsequently, a
number of parallel deep learning frameworks have been proposed for training or
running inference on MoEs on multi-GPU clusters.  These frameworks usually
combine the aforementioned parallel deep learning algorithms with expert
parallelism, which entails computing expert blocks in an embarrassingly
parallel manner on multiple GPUs. Rajbhandari et al.~present
DeepSpeed-MoE~\cite{ds-moe-systems}, a state-of-the-art system for training and
running inference on MoEs that combines expert parallelism with ZeRO's data
parallelism. Nie et al.~combine develop highly optimized kernels for routing
and all-to-all communication in their framework called HetuMoE~\cite{hetu-moe}.
In SE-MoE, the authors combine expert parallelism with out-of-core training,
wherein they store model data on the CPU memory and SSDs to enable training of
extremely large MoEs~\cite{se-moe}.  Artetxe et al.~employ Pytorch's Fully
Sharded Data Parallelism (FSDP~\cite{pytorchFSDP}) to training MoEs with
trillions of parameters~\cite{fb-moe}. In their framework called Tutel, Hwang
et al.~propose several optimizations for training MoEs at scale such as
optimized kernels for the routing function, an efficient 2D hierarchical
algorithm for all-to-all communication, and adaptive parallelism for dynamic
MoE workloads~\cite{hwang2022tutel}.

\section{Conclusion}
Deep learning researchers have recently started exploring Mixture-of-Experts
(MoE) to combat the increasing computational demands of large neural networks.
Prior state-of-the-art for parallelizing MoE architectures combined data and
expert parallelism but not tensor parallelism.  In this work, we presented a
novel, hybrid parallel algorithm that combines tensor, expert, and data
parallelism to enable the training of MoE models with $4-8 \times$ larger base
models than the current state-of-the-art, DeepSpeed-MoE. We identified an
abnormal memory spike in the optimizer that only occurs for MoEs and proposed a
tiled implementation of the optimizer to alleviate this problem. We also showed
that a naive combination of tensor and expert parallelism results in
significant redundancy in collective communication, and proposed communication
optimizations to solve this issue.  Finally, we conducted a thorough set of
empirical experiments to validate the effectiveness of our proposed framework.
Future work involves the addition of pipelining as a new dimension of
parallelism in order to scale our framework to base models that cannot fit on a
single node.

\begin{acks}
This work was supported by funding provided by the University of Maryland
College Park Foundation. This research used resources of the Oak Ridge
Leadership Computing Facility at the Oak Ridge National Laboratory, which is
supported by the Office of Science of the U.S.~Department of Energy (DOE) under
Contract No.~DE-AC05-00OR22725. This research also used resources of the
Argonne Leadership Computing Facility, which is a DOE Office of Science User
Facility supported under Contract No.~DE-AC02-06CH11357.

\end{acks}

\bibliographystyle{ACM-Reference-Format}
\bibliography{./bib/cite,./bib/pssg}

\balance

\end{document}